\documentclass[11pt]{article}

\usepackage{acl}

\usepackage{times}
\usepackage{latexsym}

\usepackage[T1]{fontenc}
\usepackage[utf8]{inputenc}
\usepackage{microtype}
\usepackage{inconsolata}
\usepackage{graphicx}
\usepackage{amsmath}
\usepackage{amssymb}

\usepackage{booktabs}      % \toprule \midrule \cmidrule \bottomrule
\usepackage{multirow}      % \shortstack in table headers
\usepackage{algorithm}     % algorithm environment (supplementary)
\usepackage{algorithmic}   % \REQUIRE \STATE \FOR etc.
\usepackage{float}         % [H] float placement (supplementary)
\usepackage{listings}      % \begin{lstlisting}[style=prompt]
\usepackage{xspace}        % \skillcdg \rawskill \lightrag macros

\newcommand{\skillcdg}{SkillCDG\xspace}
\newcommand{\rawskill}{RawSkill\xspace}
\newcommand{\lightrag}{LightRAG\xspace}
\newcommand{\code}[1]{\texttt{#1}}

\lstdefinestyle{prompt}{
  basicstyle=\footnotesize\ttfamily,
  frame=none,
  xleftmargin=1em,
  breaklines=true,
  breakatwhitespace=false,
  columns=fullflexible,
  keepspaces=true,
  showstringspaces=false,
  aboveskip=0.5em,
  belowskip=0.5em
}

\title{Long SKILL Compliance as Logical Reasoning: Closure-Grounded Detection with Scaling-Guided On-Policy Distillation}

\author{%
  Shuaitao Zhao \thanks{Equal contribution.} \quad Feng Ni \footnotemark[1] \quad
  Lichao Ma \quad \\
  \textbf{Jiaye Lin} \quad 
  \textbf{Fei Han} \quad 
  \textbf{Yang Wei} \quad
  \textbf{Lu Pan} \quad \\
  {Meituan, Beijing, China } \\
  \texttt{\{zhaoshuaitao,nifeng02\}@meituan.com} \\
}

\begin{document}
\maketitle

\begin{abstract}
The increasing complexity of enterprise business scenarios has promoted the widespread adoption of long SKILL documents in agent systems, posing new challenges for compliance detection: large models incur substantial inference costs, while small models may fail to maintain detection accuracy.
To address this gap, we propose SkillCDG, a graph-based framework for long SKILL compliance detection.
SkillCDG represents complex business policies as a two-layer constraint dependency graph, where the upper layer indexes SKILL descriptions for scenario routing and the lower layer captures dependencies among atomic constraints within each SKILL.
During inference, two-level retrieval followed by dependency closure supports compliance judgment and source traceability.
We comprehensively evaluate the framework on three enterprise datasets and two controlled public benchmark variants.
Experimental results demonstrate that SkillCDG outperforms baseline methods by up to 12.8 percentage points in detection F1 score, while reducing token consumption by a maximum 64.3\%.
Moreover, we further investigate the inherent relationships among policy-graph complexity, model scale, and detection performance.
Comparative experiments conducted on four checkpoints from a single model family validate a concise and effective scaling trend: end-to-end detection correctness exhibits a complexity-differentiated scaling pattern, and the complexity metric derived from the constraint dependency graph can effectively quantify instance difficulty and the performance improvement potential of models.
Leveraging this insightful scaling trend, we conduct adaptive training sample selection and adopt on-policy distillation to efficiently enhance the compliance detection capability of small-scale models.
\end{abstract}

\section{Introduction}
\label{sec:introduction}

Large language model (LLM) agents increasingly rely on \emph{SKILLs} to
encapsulate domain knowledge, operational procedures, and business policies.
Recent work evaluates whether SKILLs improve task performance and are followed
faithfully by agents
\citep{li2026skillsbench,han2026sweskillsbench,zhong2026skilllearnbench,
gao2026skillaudit,wang2026skilltester}.
Enterprise SKILLs, however, are often long documents containing workflows,
eligibility conditions, mandatory checks, and exceptions. Besides guiding
execution, they serve as the specification for post-hoc auditing
\citep{balaji2026beyond,yang2026complibench}. Given an interaction and an agent
response, an auditor must determine whether every applicable constraint was
satisfied and, for a violation, identify both the violated rule and its
supporting interaction evidence.

\begin{figure}[htbp]
  \centering
  \includegraphics[width=\columnwidth]{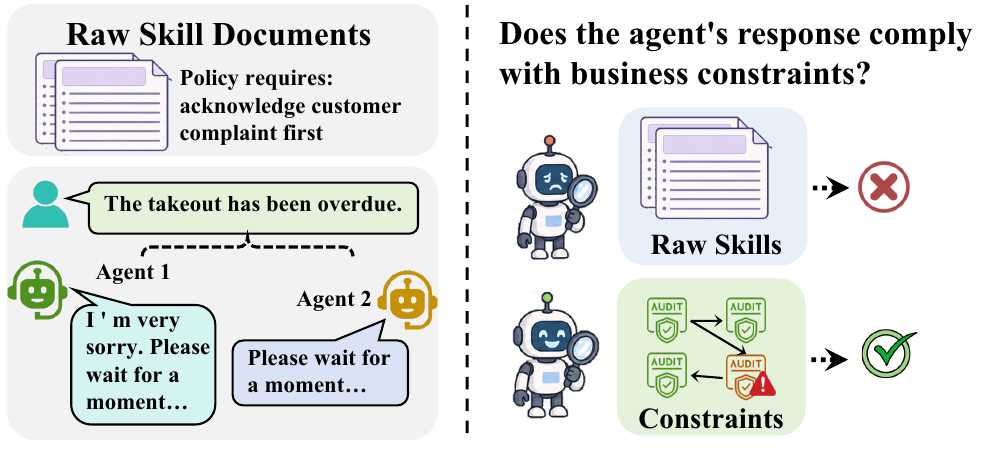}
  \caption{Identical responses may result from compliant and non-compliant
  procedures. Reliable auditing requires the applicable rule and its complete
  set of policy prerequisites.}
  \label{fig:teaser}
\end{figure}

Auditing differs fundamentally from execution. Execution starts from a
business state that activates a procedure and leads to an action. Auditing
starts from the observed action and reasons backward to recover the governing
workflow, active constraints, and mandatory prerequisites. As illustrated in Figure~\ref{fig:teaser}, two agents may issue
the same refund, for example, although only one verified eligibility, obtained
confirmation, and respected the applicable limit. The outcome alone cannot
distinguish them; compliance depends on the complete policy-governed path
\citep{yao2024tau,balaji2026beyond,yang2026complibench}.

Instruction-following benchmarks test explicit and compositional constraints
\citep{zhou2023instruction,he2024multi,
pyatkin2025generalizingverifiableinstructionfollowing,wen2024benchmarking,
wu2024lifbench,diao2025guidebench},
while interactive benchmarks evaluate policy adherence during task completion
\citep{yao2024tau,barres2025tau,balaji2026beyond}. CompliBench directly tests
whether LLM judges detect and localize violations
\citep{yang2026complibench}. These evaluations commonly expose policies as flat
documents or rule collections. The judge must retrieve relevant rules,
reconstruct their prerequisites, and apply the resulting obligations
simultaneously. An incorrect verdict may therefore arise from failed retrieval,
dependency inference, or rule application, but end-to-end accuracy cannot
separate these causes. This coupling also obscures what a stronger judge
actually improves: policy access, logical reconstruction, or the final
comparison between obligations and observed behavior.

Long-context and retrieval methods only partially address this coupling.
Nominally long context windows do not ensure robust use of dispersed evidence
\citep{liu2024lost,bai2024longbench,du2025context}, while compression can
discard task-relevant details
\citep{jiang2023longllmlingua,pan2024llmlingua2}. Retrieval-augmented
generation supplies compact external evidence \citep{lewis2020rag}, and
graph-based methods improve multi-hop access through semantic relations
\citep{gutierrez2024hipporag,edge2024graphrag,guo2024lightrag}. Compliance,
however, requires mandatory logical dependencies: omitting a prerequisite or
exception may reverse the verdict. General-purpose semantic graphs do not
guarantee such dependency completeness. Difficulty therefore depends not only
on document length, but also on how relevant rules and their prerequisites are
distributed throughout the policy.

Long SKILL auditing thus poses three coupled challenges: scenario
routing identifies the governing SKILL and workflow; dependency
completion recovers all activated prerequisites, exceptions, and checks; and
audit traceability retains links between obligations and their source
evidence. Flat policy representations leave all three implicit inside the
judge.

We propose \textbf{SkillCDG}, which formulates long SKILL compliance detection
as action-conditioned policy reasoning. SkillCDG converts enterprise SKILL format
policies into a two-layer constraint dependency graph. The upper layer
indexes SKILL metadata, e.g., name and description, to locate the governing process.
The lower layer decomposes each SKILL into
atomic condition-action constraints linked to their source spans. A typed
\texttt{require} edge denotes a mandatory
prerequisite for auditing another rule, making both rule applicability and
triggered obligations explicit.

During inference, SkillCDG first retrieves candidate scenarios using the
interaction and agent response. It then retrieves atomic seed rules within the
selected SKILLs and expands their \texttt{require} dependencies into an
instance-specific closure. This compact subgraph contains both directly
relevant rules and every prerequisite reachable in the extracted CDG. An LLM
judge uses the closure to predict compliance, while retained source spans provide
auxiliary evidence for inspecting violations. The pipeline therefore separates policy retrieval,
dependency completion, and judgment while retaining links from every atomic
constraint to its original policy span.

The structured representation further enables us to ask whether larger judges
continue to justify their cost as policies become denser and more
interdependent. Scaling-law studies relate performance to model size, data,
compute, and context properties
\citep{kaplan2020scaling,hoffmann2022computeoptimal,snell2025scaling,
yue2025inference,montgomery2025contextaware},
but do not isolate business-policy structure. We characterize this structure
with \emph{rule density}, the proportion of policy tokens retained in auditable atomic rules,
and \emph{dependency coupling}, the number of prerequisite edges relative to
the number of constraints. The two factors distinguish policies with many
independent rules from tightly coupled dependency networks.

We evaluate SkillCDG on three enterprise datasets and two controlled public
benchmark variants. Within the Qwen3.5 family, we hold the SkillCDG
representation and inference settings fixed across four checkpoints, isolating
the relationship between model scale and sample-level policy-graph complexity.
The analysis reveals a useful decomposition: end-to-end capacity improves and
then saturates with model scale, while CDG-derived complexity represents
residual instance difficulty across checkpoints. The data favor this
parsimonious form over an additional scale--complexity attenuation
interaction. We use the selected predictor to prioritize instances with
greater teacher--student headroom for scaling-guided on-policy distillation
(OPD), linking predictive model selection to cost-aware small-model training
\citep{gu2024minillm,agarwal2024gkd}.

Our main contributions are:
\begin{itemize}
  \item We formulate long SKILL auditing as action-conditioned policy reasoning
  and introduce a two-layer CDG that supports SKILL-level scenario routing and atomic
  prerequisite dependencies.
  \item We develop two-level retrieval and dependency closure to construct
  compact policy contexts for compliance judgment, with source spans retained
  as auxiliary audit evidence.
  \item Across three enterprise datasets, two public-benchmark variants, and
  multiple model scales, we estimate and validate a saturating empirical
  relationship between model scale, policy-graph complexity, and correctness.
  \item We translate the fitted teacher--student headroom into scaling-guided
  OPD sample priorities under a fixed distillation budget.
\end{itemize}

\section{Related Work}
\label{sec:related_work}

\subsection{Instruction Following and Agent Compliance}
Instruction tuning and human-feedback alignment improve general instruction
adherence \citep{ouyang2022training}, motivating evaluations of explicit,
multilingual, multi-turn, verifiable, and long-context constraints
\citep{zhou2023instruction,he2024multi,he2024can,
pyatkin2025generalizingverifiableinstructionfollowing,wen2024benchmarking,
wu2024lifbench}. Agent-focused benchmarks extend this setting to domain
guidelines and tool-mediated tasks
\citep{diao2025guidebench,qi2025agentif,yao2024tau,barres2025tau,
balaji2026beyond}. CompliBench further evaluates whether an LLM judge can
detect and localize policy violations \citep{yang2026complibench}. These works
measure end-to-end adherence, but generally leave rule retrieval and
prerequisite reconstruction implicit inside the evaluated model.

\subsection{Agent Skills and Structured Policy Access}
Recent benchmarks study the utility, generation, security, and retrieval of
Agent Skills \citep{li2026skillsbench,han2026sweskillsbench,
zhong2026skilllearnbench,
gao2026skillaudit,wang2026skilltester,
su2026skillretrieval}. Long-context evaluations reveal that larger context
windows alone do not guarantee reliable evidence use
\citep{liu2024lost,bai2024longbench,du2025context}. Compression can reduce
irrelevant input \citep{jiang2023longllmlingua,pan2024llmlingua2}, and RAG
augments generation with retrieved external evidence \citep{lewis2020rag}.
Graph-based retrieval and automatic knowledge-graph construction further
improve multi-hop access through semantic relations
\citep{gutierrez2024hipporag,edge2024graphrag,guo2024lightrag,mo2025kggen}.
SkillCDG instead represents mandatory dependencies among auditable
condition-action constraints, so graph traversal constructs a prerequisite
closure rather than a set of semantically related passages.

\subsection{Scaling and Distillation}
Classical scaling laws relate loss to model size, data, and compute
\citep{kaplan2020scaling,hoffmann2022computeoptimal}; later work studies
test-time compute, retrieval-augmented inference, and context-dependent scaling
\citep{snell2025scaling,yue2025inference,montgomery2025contextaware}. These
formulations do not model policy structure as a task variable. We estimate an
empirical, task-specific relationship between model size, policy-graph
complexity, and compliance correctness. Knowledge distillation transfers
teacher behavior to smaller models \citep{hinton2015distilling}. Conventional
offline distillation learns from fixed reference or teacher-generated
sequences, whereas on-policy distillation evaluates student-generated
trajectories under the teacher, reducing the mismatch between training prefixes
and those visited during autoregressive inference
\citep{gu2024minillm,agarwal2024gkd}. Our downstream distillation study uses predicted
teacher--student headroom to allocate a fixed on-policy training budget.

\section{Methodology}
\label{sec:method}

\begin{figure*}[htbp]
  \centering
  \includegraphics[width=\textwidth]{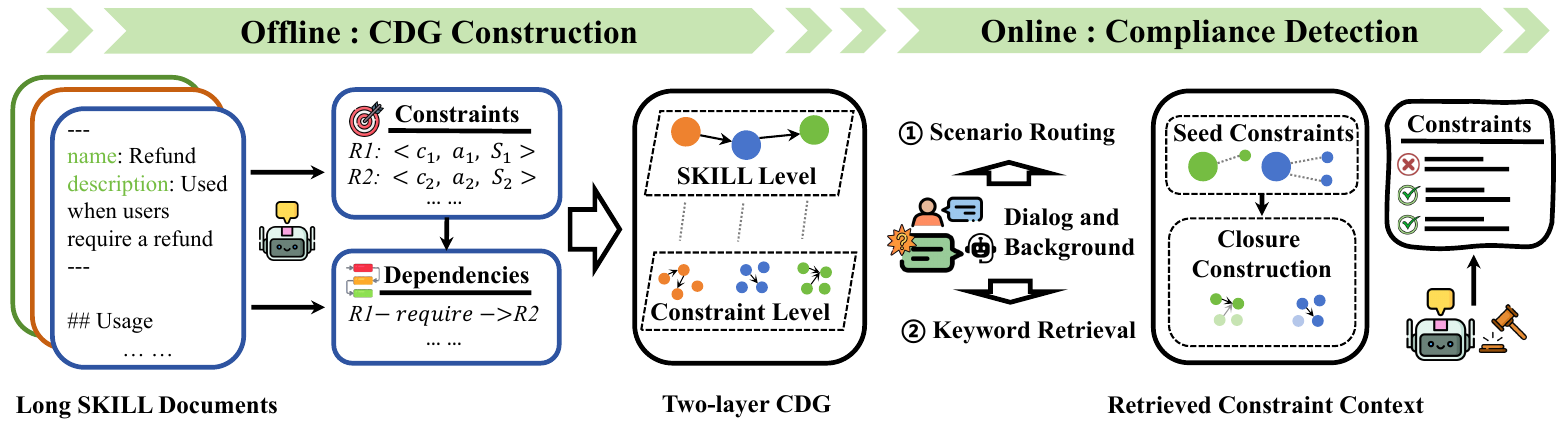}
  \caption{Overview of SkillCDG. The offline stage converts long SKILL business policies into a two-layer constraint dependency graph. The online stage performs SKILL-level retrieval, constraint-level retrieval, dependency closure, and LLM-based compliance judgment.}
  \label{fig:overview}
\end{figure*}

Our methodology has two components. First, SkillCDG represents long SKILL
business policies as a Constraint Dependency Graph (CDG) and retrieves a
graph-complete rule context for compliance judgment. Second, a
complexity-aware empirical scaling model jointly characterizes model capacity
and policy-graph difficulty. The former removes redundant input while restoring
prerequisites; the latter supports model and sample selection. Each policy document is a
\emph{SKILL}, denoted $D_j$, and the collection is
$\mathcal D=\{D_1,\ldots,D_m\}$. As illustrated in
Figure~\ref{fig:overview}, SkillCDG contains an offline graph-construction stage
and an online retrieval-and-judgment stage.

\subsection{SkillCDG: Constraint Dependency Graph-Based Compliance Detection}
\label{sec:skillcdg}

\subsubsection{Problem Formulation}
\label{sec:problem_formulation}

Given a conversation history $H$, current environment state $O$, the SKILL collection $\mathcal D$, and an agent response $A$, the task of policy compliance detection is to judge whether $A$ violates policies in $\mathcal D$, formulated as:
\begin{equation}
  \hat y=f_\theta(H,O,\mathcal D,A),\quad \hat y\in\{0,1\},
  \label{eq:task}
\end{equation}
where $\hat y$ is the predicted label. The gold label $y$ follows the same
encoding: $y=1$ indicates a violation and $y=0$ indicates compliance.
The interaction state is essential because the same action can be compliant
under one condition and prohibited under another; the task therefore depends
on policy applicability rather than surface similarity.

Each SKILL policy can be decomposed into a set of \emph{atomic constraint rules}. Each rule is defined as a triple $r_i=\langle c_i,a_i,\mathcal S_i\rangle$, where $c_i$ specifies the applicable condition, $a_i$ defines the normative agent action (required, prohibited, or permitted), and $\mathcal S_i$ retains the original policy text spans for audit traceability. Rules are not fully independent: evaluating one rule may require the compliance status of other rules as a logical prerequisite, which motivates our graph-based dependency modeling.
Atomic decomposition separates applicability from the required action, and
source spans identify both the evaluated obligation and its policy origin.

\subsubsection{Offline Stage: Constraint Dependency Graph Construction}
\label{sec:cdg_construction}

We first formalize the graph structure and dependency semantics, then describe the three-step construction pipeline.

\textbf{Formal Definitions}. For a SKILL $D_j$ with rule set $\mathcal{R}_j$, we define a directed \textit{require} relation: an edge $(r_i, r_k) \in \mathcal{E}_j$ (denoted $r_i \rightarrow r_k$) holds if and only if determining the compliance status of $r_i$ necessarily requires evaluating $r_k$ as a prerequisite. The rule set $\mathcal{R}_j$ and edge set $\mathcal{E}_j$ form the rule-level dependency graph $\mathcal{G}_j = (\mathcal{R}_j, \mathcal{E}_j)$. This mandatory semantics distinguishes prerequisites from merely related concepts.

\textbf{Block-wise Atomic Constraint Extraction}. Long SKILL documents are split into semantic blocks along Markdown heading boundaries. For each block, an LLM extracts atomic rules and identifies require dependencies within the block, forming a local subgraph. This keeps extraction focused on a coherent segment and reduces distraction from distant text.

\textbf{Cross-Block Graph Fusion}. Local subgraphs are incrementally merged into the complete rule graph $\mathcal{G}_j$ for each SKILL. During fusion, semantically equivalent rules across blocks are deduplicated, their source spans are unified, and cross-block require edges are supplemented. Fusion repairs block-level fragmentation by reconnecting prerequisites across sections without duplicating obligations.

\textbf{Two-Layer Graph Architecture}. The full CDG $\mathcal{G}^{\mathrm{CDG}}$ consists of two hierarchical layers. The upper layer is a SKILL-level routing index whose nodes retain SKILL metadata for scenario selection. The lower layer contains the atomic rule graphs $\{\mathcal{G}_j\}_{j=1}^{m}$ for individual SKILLs. This separation excludes unrelated scenarios while retaining fine-grained dependencies within each selected policy.

Unlike entity-centric knowledge graphs built for general retrieval with heterogeneous relations \citep{edge2024graphrag,guo2024lightrag}, SkillCDG assigns a uniform \textit{require} semantics to all rule-level edges, eliminating relation-type heterogeneity. Graph traversal in each selected SKILL computes the transitive prerequisite closure for compliance judgment. This design brings two core properties: (1) \textit{Source traceability}: every rule retains direct links to its original policy text; (2) \textit{Unified audit semantics}: rule-graph operations directly serve the completeness requirement of compliance decision-making.

\subsubsection{Online Stage: Compliance Detection Pipeline}
\label{sec:compliance_pipeline}

The online pipeline maps the current interaction $q=(H,O,A)$ to a compact,
graph-complete policy context via three sequential steps.

\begin{itemize}
  \item[(i)] \textbf{Two-Level Policy Retrieval}. Audit keywords are extracted from the interaction context for coarse-to-fine retrieval. At the SKILL level, keywords are matched against SKILL names and descriptions to remove irrelevant scenarios. Within each selected SKILL, keywords are matched against rule conditions, actions, and source spans to locate seed obligations and limit the judgment context.
  \item[(ii)] \textbf{Dependency Closure Expansion}. Starting from the seed rule set, we traverse require edges in $\mathcal{G}_j$ to compute the full dependency closure. Retrieval alone may find a target rule but omit an eligibility check or exception needed to interpret it. Closure repairs this incompleteness by retaining every reachable prerequisite while still filtering unrelated rules.
  \item[(iii)] \textbf{Compliance Judgment}. The rule closure is converted into structured policy text and fed into an LLM together with the interaction context. Explicit conditions, actions, and prerequisites let the judge apply rules without reconstructing the full policy.
\end{itemize}

Two-level retrieval narrows the policy scope to be judged, and dependency closure yields complete judgment basis based on the \texttt{require} edges in the extracted CDG. Filtering irrelevant redundant text further reduces model inference overhead, bringing lower inference latency and higher detection accuracy. Detailed examples and human evaluation results are included in the supplementary materials.

\subsection{Complexity-Aware Empirical Scaling Trend}
\label{sec:complexity-scaling}

The CDG also makes policy structure measurable. For instance $i$, let
$\mathcal D_i\subseteq\mathcal D$ denote its associated SKILL documents and
$\mathcal G_i=(\mathcal R_i,\mathcal E_i)$ denote their induced rule graph,
where $\mathcal R_i$ contains atomic rules and $\mathcal E_i$ contains their
\texttt{require} dependencies. We define rule density, dependency coupling,
and their composite policy-graph complexity as
\begin{equation}
I_{d,i}=\frac{T(\mathcal R_i)}{T(\mathcal D_i)},\qquad
I_{c,i}=\frac{|\mathcal E_i|}{|\mathcal R_i|},\qquad
C_i=I_{d,i}I_{c,i},
\label{eq:graph-complexity}
\end{equation}
where $T(\cdot)$ counts tokens, including all documents when its argument is
a set. Thus, high $C_i$
indicates that constraints are both dense in the document and strongly
coupled through prerequisites.
Density measures auditable content, while coupling measures prerequisite
reasoning; their product distinguishes long sparse documents from compact,
dependency-heavy policies.

Let $N$ be the parameter count and
$Z_{i,N}=\mathbf{1}[\hat y_{i,N}=y_i]$ indicate whether a model with parameter
count $N$ correctly judges instance $i$. We use a flexible candidate in which correctness
can saturate with scale and the scaling trajectory can vary with policy-graph
complexity. Following context-aware scaling formulations
\citep{montgomery2025contextaware}, a mechanism-driven candidate expresses
this behavior as
\begin{equation}
\begin{array}{rcl}
\mathrm{logit}\,P(Z_{i,N}=1)
&=&\gamma_{\mathrm{data}(i)}-\eta\widetilde C_i\\
&&+\beta\!\left(1-e^{-\rho\widetilde N^\alpha}\right)
e^{-\lambda\widetilde C_i},
\end{array}
\label{eq:complexity-scaling}
\end{equation}

where $\widetilde N=N/1\mathrm{B}$,
$\widetilde C_i=C_i/C_0$, $C_0$ is the training-fold median, and
$\gamma_{\mathrm{data}(i)}$ is a dataset intercept; the remaining parameters
satisfy $\beta,\rho,\alpha>0$, while $\eta$ and $\lambda$ remain unconstrained so
that the data, rather than the parameterization, determine the direction of
the complexity effects. The first exponential term models saturation with
scale, while the interaction term allows complexity to alter scale-dependent
gains. To avoid imposing a particular interaction, we compare
Eq.~\ref{eq:complexity-scaling}
with an additive logistic model and a nested saturating model that fixes
$\lambda=0$.

Let $\widehat P_b(C,d)=\widehat P(Z=1\mid N=b,C,\mathrm{data}=d)$ and let
$Q_C(p)$ denote the empirical $p$-quantile of policy-graph complexity. For
parameter count $b$ and dataset $d$, we measure the gain over the 2B checkpoint
and the change in this gain between high- and low-complexity policies:
\begin{equation}
\begin{array}{rcl}
G_b(C,d) & = & \widehat P_b(C,d)-\widehat P_{2\mathrm{B}}(C,d),\\
\Delta G_b(d) & = & G_b(Q_C(0.9),d)-G_b(Q_C(0.1),d).
\end{array}
\label{eq:scaling-gain}
\end{equation}

The gain statistics locate where additional capacity is useful, supporting
model choice and sample prioritization. Models are estimated from sample-level Bernoulli
outcomes and selected using session-grouped cross-validation with the
one-standard-error rule. Parameters are fitted by global differential
evolution followed by local optimization. Session-clustered bootstrap yields
confidence intervals, and leave-one-scale-out evaluation tests whether the
relationship predicts an unseen checkpoint. Grouped validation prevents
leakage across related interactions, while the latter test measures
generalization beyond fitted scales.

\section{Experiments}
\label{sec:experiments}

We organize the evaluation around four research questions:
\begin{itemize}
    \item \textbf{RQ1:} How effectively and efficiently does SkillCDG detect
    violations in long SKILL documents?
    \item \textbf{RQ2:} Can model scale and CDG-derived policy complexity
    jointly predict compliance-detection performance?
    \item \textbf{RQ3:} What are the respective contributions of keyword
    retrieval and dependency closure?
    \item \textbf{RQ4:} Can scaling-derived headroom improve the allocation of
    a fixed on-policy distillation budget?
\end{itemize}

\subsection{Experimental Setup}
\label{sec:exp-setup}

\paragraph{Datasets.}
We evaluate practical utility on three anonymized enterprise datasets and
out-of-domain transfer on one public benchmark and its controlled long-policy
variant. Anonymized enterprise datasets include \textbf{Fulfillment}, \textbf{AfterSales} and \textbf{MerchantSupport}.
\textbf{Fulfillment} covers consumer-facing order-fulfillment conversations
governed by long business policies. \textbf{AfterSales} contains post-order
service conversations organized around standardized business workflows.
\textbf{MerchantSupport} covers merchant-facing consultation and operational
support. \textbf{CompliBench} evaluates violation detection under
normal-length enterprise guidelines \citep{yang2026complibench}.
\textbf{CompliBench-Long} preserves the original conversations and labels but
augments each policy with irrelevant intents and redundant paraphrases,
creating a controlled long-policy setting with substantial distractor text.
Table~\ref{tab:exp-data} summarizes the five evaluation settings.

\begin{table}[htbp]
\centering
\small
\setlength{\tabcolsep}{2.2pt}
\begin{tabular}{lrrr}
\toprule
Dataset & Samples  & \# SKILLs & Avg.\ tokens \\
\midrule
Fulfillment      & 1,442 & 14 & 10.6k \\
AfterSales       & 1,536 &  12 &  7.0k \\
MerchantSupport  &   870 & 32 &  5.0k \\
CompliBench      & 2,710 & 86 &    0.7k \\
CompliBench-Long & 2,710 & 86 &  9.5k \\
\bottomrule
\end{tabular}
\caption{Dataset statistics. Average tokens is measured by Qwen3.5 series tokenizer.}
\label{tab:exp-data}
\end{table}

\paragraph{Baseline methods.}
We compare SkillCDG with two representative policy-access strategies.
RawSkill supplies all associated SKILL documents directly to the
judge. LightRAG builds an entity-relation graph and retrieves policy
context in its keyword configuration, without an additional embedding model.
SkillCDG retrieves SKILLs and seed rules by keyword and then computes
the transitive closure of their \texttt{require} dependencies. 

\paragraph{Evaluation metrics.}
We report accuracy (ACC) for overall correctness and F1 score for balanced detection.

\paragraph{Implementation Details.}
We conduct experiments with open-source models including the Qwen3.5\cite{qwen35blog} series and DeepSeek-V4\cite{deepseekv4tech} series, and further perform evaluations on the closed-source GPT-5. 
The chunk length is set to 12000 for chunked processing, and Qwen3-0.6b is adopted as the embedding model. For compliance detection, the temperature parameter is fixed to 0.01 across all LLM inference calls.

\subsection{RQ1: Effectiveness and Inference Cost}
\label{sec:rq1}

Long SKILL documents contain both essential dependencies and scenario-irrelevant content.
SkillCDG performs two-stage filtering via the construction of Constraint Dependency Graphs to precisely retrieve scenario-relevant constraint rules. To evaluate the effectiveness of SkillCDG, we conduct comparisons against existing baselines across three model families and five datasets. We measure compliance detection accuracy and F1 score for all methods, and the experimental results are presented in Table~\ref{tab:rq1-effect}.

\begin{table}[t]
\small
\centering
\setlength{\tabcolsep}{5pt}
\begin{tabular}{l*{3}{cc}}
\toprule
Method
& \multicolumn{2}{c}{Fulfillment}
& \multicolumn{2}{c}{AfterSales}
& \multicolumn{2}{c}{\shortstack{Merchant\\Support}} \\
\cmidrule(lr){2-3}\cmidrule(lr){4-5}\cmidrule(lr){6-7}
& ACC & F1 & ACC & F1 & ACC & F1 \\
\midrule
\multicolumn{7}{l}{\textbf{Qwen3.5-4B}} \\
\quad RawSkill & 42.8 & 41.5 & 48.0 & 44.7 & 46.1 & 46.1 \\
\quad LightRAG & 44.6 & 40.3 & 49.6 & 45.2 & 47.3 & 48.5 \\
\quad SkillCDG & \textbf{54.4} & \textbf{54.3} & \textbf{50.6} & \textbf{50.6} & \textbf{56.0} & \textbf{52.9} \\
\midrule
\multicolumn{7}{l}{\textbf{Qwen3.5-9B}} \\
\quad RawSkill & 41.3 & 39.2 & 48.4 & 45.3 & 44.9 & 44.9 \\
\quad LightRAG & \textbf{51.2} & 48.7 & 48.3 & 49.5 & 45.7 & 45.4 \\
\quad SkillCDG & 50.0 & \textbf{50.0} & \textbf{54.8} & \textbf{54.8} & \textbf{50.8} & \textbf{50.0} \\
\midrule
\multicolumn{7}{l}{\textbf{Qwen3.5-27B}} \\
\quad RawSkill & 54.4 & 54.1 & 56.5 & \textbf{56.2} & 62.5 & \textbf{58.7} \\
\quad LightRAG & 60.3 & \textbf{54.2} & 56.2 & 49.3 & 69.1 & 45.8 \\
\quad SkillCDG & \textbf{67.6} & 46.5 & \textbf{58.3} & 40.7 & \textbf{71.7} & 45.6 \\
\midrule
\multicolumn{7}{l}{\textbf{DeepSeek-V4-Flash}} \\
\quad RawSkill & \textbf{67.9} & 57.0 & \textbf{61.1} & \textbf{57.3} & 70.5 & 46.1 \\
\quad LightRAG & 65.9 & 43.5 & 58.8 & 41.3 & \textbf{72.0} & 43.4 \\
\quad SkillCDG & 65.0 & \textbf{59.1} & 56.8 & 54.8 & 67.4 & \textbf{51.4} \\
\midrule
\multicolumn{7}{l}{\textbf{DeepSeek-V4-Pro}} \\
\quad RawSkill & \textbf{68.7} & \textbf{62.6} & \textbf{70.4} & \textbf{69.5} & \textbf{72.0} & 60.5 \\
\quad LightRAG & 67.4 & 55.6 & \textbf{70.4} & 69.3 & 71.3 & \textbf{60.8} \\
\quad SkillCDG & 60.3 & 55.6 & 60.4 & 67.8 & 68.5 & 59.5 \\
\midrule
\multicolumn{7}{l}{\textbf{GPT-5.4}} \\
\quad RawSkill & \textbf{69.9} & \textbf{64.3} & \textbf{67.8} & \textbf{66.8} & \textbf{71.0} & 56.9 \\
\quad LightRAG & 64.1 & 46.7 & 57.4 & 46.3 & 68.7 & 52.2 \\
\quad SkillCDG & 61.8 & 58.0 & 63.7 & 63.2 & 69.8 & \textbf{58.8} \\
\bottomrule
\end{tabular}
\caption{Enterprise effectiveness. ACC and F1 are reported in
percentage points. Bold indicates the best result within each model-dataset block.}
\label{tab:rq1-effect}
\end{table}

Gains are strongest for compact judges. Against RawSkill, SkillCDG improves
Qwen3.5-4B F1 by 12.8, 5.9, and 6.8 points and Qwen3.5-9B F1 by 10.8, 9.5,
and 5.1 points across the three datasets. At 27B, it further raises ACC by
13.2, 1.8, and 9.2 points. These results show that structured policy access is
most beneficial to compact judges, which are more sensitive to redundant text
and missing logical links, and remains effective at larger scales. The pattern
also shows that increasing model size does not replace policy organization:
better inputs and stronger judges address complementary parts of the task.

\paragraph{Cost evaluation.}
To assess computational overhead, we measure token consumption of SkillCDG in compliance detection. We benchmark against the raw long-text approach on three datasets and record the average per-sample token usage, as reported in Table~\ref{tab:rq1-cost-compact}. SkillCDG cuts average online token consumption by 64.3\%, 33.9\%, and 39.1\% over the three enterprise datasets by filtering out irrelevant rules while retaining dependencies. The Fulfillment dataset, featuring the longest average policy texts, yields the largest token savings, confirming that structural filtering becomes more valuable with longer rule documents.

\begin{table}[htbp]
\small
\centering
\setlength{\tabcolsep}{5pt}
\begin{tabular}{lrrr}
\toprule
Method & Fulfillment & AfterSales & \shortstack{Merchant\\Support} \\
\midrule
RawSkill & 35.0k & 10.1k & 16.8k \\
SkillCDG & 12.5k & 6.7k & 10.3k \\
\bottomrule
\end{tabular}
\caption{Mean online token usage.}
\label{tab:rq1-cost-compact}
\end{table}

\paragraph{Public-benchmark transfer.}
Without dataset-specific retuning, we compare RawSkill and SkillCDG on
CompliBench and CompliBench-Long in
Table~\ref{tab:rq1-generalization}.

\begin{table}[htbp]
\centering
\small
\setlength{\tabcolsep}{5pt}
\begin{tabular}{l*{2}{cc}}
\toprule
Variant
& \multicolumn{2}{c}{CompliBench}
& \multicolumn{2}{c}{CompliBench-Long} \\
\cmidrule(lr){2-3}\cmidrule(lr){4-5}
& ACC & F1 & ACC & F1 \\
\midrule
\multicolumn{5}{l}{\textbf{Qwen3.5-27B}} \\
\quad RawSkill & 20.4 & 25.1 & 26.3 & 27.0 \\
\quad SkillCDG & \textbf{24.2} & \textbf{25.5} & \textbf{27.8} & \textbf{29.1} \\
\midrule
\multicolumn{5}{l}{\textbf{DeepSeek-V4-Flash}} \\
\quad RawSkill & 20.8 & 20.7 & 22.6 & 22.2 \\
\quad SkillCDG & \textbf{24.5} & \textbf{24.3} & \textbf{26.4} & \textbf{25.8} \\
\bottomrule
\end{tabular}
\caption{Transfer to a public compliance benchmark and its controlled
long-policy variant.}
\label{tab:rq1-generalization}
\end{table}

SkillCDG improves ACC/F1 by $3.8/0.4$ and $1.5/2.1$ points for Qwen3.5-27B,
and by $3.7/3.6$ and $3.8/3.6$ points for DeepSeek-V4-Flash, demonstrating
consistent transfer across normal and long-policy settings. Improvements on
CompliBench-Long are especially informative because its labels and
conversations are unchanged while distractor text is added; the gains therefore
reflect resistance to irrelevant policy content rather than a change in task
semantics.

RawSkill exposes every rule and forces the judge to separate obligations from
large amounts of irrelevant text. LightRAG reduces this burden through lexical
retrieval, but semantic relevance alone does not ensure that prerequisite
rules are present. SkillCDG combines filtering with dependency closure:
filtering controls context length, while closure preserves the constraint chain
needed for compliance reasoning. This explains the joint improvement in
effectiveness and token cost.

\subsection{RQ2: Complexity-Aware Empirical Scaling}
\label{sec:rq2}

\begin{figure}[htbp]
\centering
\includegraphics[width=0.8\columnwidth,height=7cm,keepaspectratio]{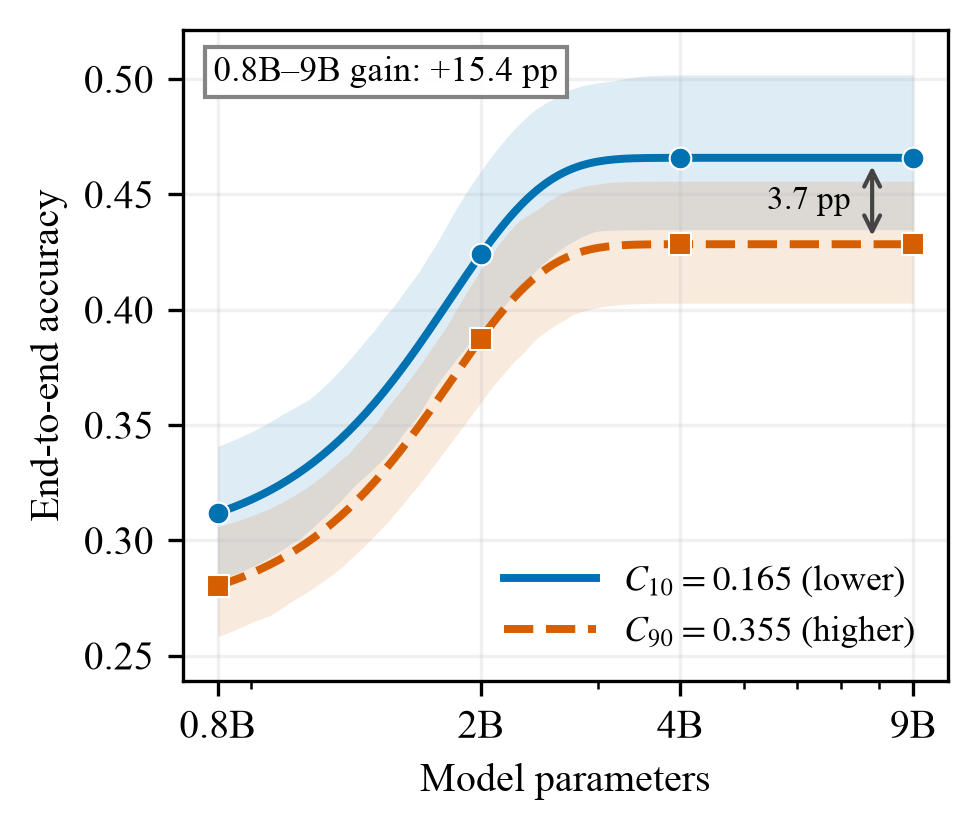}
\caption{Fitted end-to-end correctness at the 10th and 90th percentiles of CDG complexity on three business datasets.}
\label{fig:rq2-scaling}
\end{figure}

We further investigate the relationship between model scale and graph complexity based on the structure of Constraint Dependency Graphs. We fit task-specific scaling equations using samples from all five datasets with model sizes \(N\in\{0.8,2,4,9\}\)B, taking model scale and instance-level policy-graph complexity as explanatory variables. Session-grouped five-fold cross-validation is adopted to select the optimal fitted model. Leave-one-scale-out evaluation is used to test the prediction capability for unseen model scales, and 1,000 session-clustered bootstrap replicates produce 95\% confidence intervals. Figure~\ref{fig:rq2-scaling} exhibits an obvious complexity-dependent pattern: the performance on low-complexity samples converges rapidly as model scale increases, while high-complexity samples are harder to judge and attain larger performance gains from larger models. The experimental results of the 27B model align with the fitted trend, verifying the extrapolative validity of the model.

The fitted trend reveals the complementary roles of the two variables: model scale characterizes the available model capacity, while CDG complexity identifies samples where extra capacity can deliver improvements. Based on this, we can estimate detection performance under candidate model sizes, avoid deploying oversized judges for saturated simple samples, and select instances with high potential for performance improvement. This insight motivates us to estimate teacher-student headroom under the unified closure representation for model distillation, which we explore in RQ4.

\subsection{RQ3: Ablation Study}
\label{sec:rq3}

To evaluate the effectiveness of individual components within SkillCDG, we perform ablation studies on its two core modules: keyword-based filtering and dependency closure computation. 
When keyword retrieval is removed, the judge is fed with all dependency rules to verify the efficacy of the filtering mechanism. Without dependency closure, only matched rules are retained, which helps validate whether complete prerequisite reasoning is indispensable.
We conduct experiments on three enterprise datasets and measure accuracy and F1 score for each ablation variant. Experimental results are illustrated in Table~\ref{tab:rq3-compact}.

\begin{table}[htbp]
\centering
\small
\setlength{\tabcolsep}{3pt}
\begin{tabular}{l*{3}{cc}}
\toprule
Variant & \multicolumn{2}{c}{Fulfillment}
& \multicolumn{2}{c}{AfterSales}
& \multicolumn{2}{c}{\shortstack{Merchant\\Support}} \\
\cmidrule(lr){2-3}\cmidrule(lr){4-5}\cmidrule(lr){6-7}
& ACC & F1 & ACC & F1 & ACC & F1 \\
\midrule
\multicolumn{7}{l}{\textbf{Qwen3.5-27B}} \\
\quad SkillCDG
& \textbf{67.6} & 46.5 & \textbf{58.3} & \textbf{40.7} & 71.7 & \textbf{45.6} \\
\quad w/o Keyword
& 66.0 & \textbf{47.4} & 56.3 & 38.7 & \textbf{72.7} & 43.6 \\
\quad w/o Closure
& 61.7 & 43.7 & 45.6 & 33.2 & 66.1 & 40.4 \\
\midrule
\multicolumn{7}{l}{\textbf{DeepSeek-V4-Flash}} \\
\quad SkillCDG
& 65.0 & \textbf{59.1} & 56.8 & 54.8 & 67.4 & \textbf{51.4} \\
\quad w/o Keyword
& \textbf{69.7} & 56.2 & \textbf{57.4} & \textbf{55.1} & \textbf{69.2} & 48.8 \\
\quad w/o Closure
& 68.7 & 53.7 & 55.6 & 53.2 & 66.1 & 48.4 \\
\bottomrule
\end{tabular}
\caption{Ablation study results.}
\label{tab:rq3-compact}
\end{table}

Experimental results demonstrate that removing either component degrades the overall performance of SkillCDG. 
Without the keyword filtering mechanism, the context length increases drastically and the inference overhead of the discriminant model rises sharply alongside degraded detection performance. This issue is more pronounced on the Qwen3.5-27B model, as it supports a shorter maximum context window compared to DeepSeek-V4-Flash. 
In contrast, removing the dependency closure computation causes more severe performance degradation, with a maximum F1 drop of 7.5 points for the Qwen3.5-27B model. 
This verifies that rules obtained solely through direct matching are insufficient for complex policy reasoning. 
The two components cooperate synergistically to address the constraint compliance challenges inherent in long SKILL scenarios. 
Dependency closure serves as the core mechanism for ensuring detection performance, recovering eligibility conditions, exception constraints, and mandatory check rules that are easily overlooked by simple lexical matching. 
Meanwhile, keyword filtering improves inference efficiency by eliminating scenario-irrelevant rules before compliance judgment. The combination of the two modules yields a compact input context while fully preserving the completeness of prerequisite condition reasoning.

\subsection{RQ4: Scaling-Guided On-Policy Distillation}
\label{sec:rq4}

Uniform OPD treats all training samples as equally valuable, even though the
scaling results show that simple samples may already be saturated for the
student. This can spend a fixed training budget on examples with little
remaining transfer value. We instead use the fitted model in
Eq.~\ref{eq:complexity-scaling} to estimate where
Qwen3.5-9B offers Qwen3.5-4B the greatest headroom. For instance $i$ from
dataset $d_i$, the score is
\begin{equation}
g_i=\widehat P_{9\mathrm{B}}(C_i,d_i)
    -\widehat P_{4\mathrm{B}}(C_i,d_i).
\label{eq:opd-score}
\end{equation}
Both methods preserve identical dataset--label quotas; Uniform OPD samples
randomly within strata, while Scaling-guided OPD selects the highest $g_i$.
Each method uses 2,000 samples with the same model architecture, optimization
schedule, and training budget; only sample selection differs. This matched
design attributes any performance difference to the allocation strategy
rather than additional data or compute.

Let $F_B=\mathrm{F1}(\mathrm{Base\mbox{-}4B})$ and
$F_T=\mathrm{F1}(\mathrm{Teacher\mbox{-}9B})$. We quantify the fraction of
the initial teacher--student gap recovered by method $M$ as
\begin{equation}
\mathrm{Recovery}(M)=
\frac{\mathrm{F1}(M)-F_B}{F_T-F_B}\times100\%.
\label{eq:gap-recovery}
\end{equation}

\begin{table}[htbp]
\centering
\small
\setlength{\tabcolsep}{1.5pt}
\begin{tabular}{l*{3}{cc}}
\toprule
Model
& \multicolumn{2}{c}{Fulfillment}
& \multicolumn{2}{c}{AfterSales}
& \multicolumn{2}{c}{\shortstack{Merchant\\Support}} \\
\cmidrule(lr){2-3}\cmidrule(lr){4-5}\cmidrule(lr){6-7}
& ACC & F1 & ACC & F1 & ACC & F1 \\
\midrule
4B Base
  & 52.4 & 50.3 & 51.6 & 52.6 & 56.0 & 52.9 \\
9B Teacher
  & 56.0 & 55.0 & \textbf{54.8} & 53.8 & \textbf{57.8} & \textbf{58.0} \\
\shortstack[l]{4B +\\Uniform OPD}
  & 54.4 & 54.3 & 50.6 & 53.6 & 56.0 & 53.9 \\
\shortstack[l]{4B +\\Scaling-guided OPD}
  & \textbf{56.4} & \textbf{56.3} & 53.6 & \textbf{54.6} & 57.3 & 54.9 \\
\bottomrule
\end{tabular}
\caption{Scaling-guided on-policy distillation with Qwen3.5-4B as the student and Qwen3.5-9B as the teacher.}
\label{tab:rq4-compact}
\end{table}

Table~\ref{tab:rq4-compact} shows that scaling guidance improves ACC/F1 by
$2.0/2.0$, $3.0/1.0$, and $1.3/1.0$ points on Fulfillment, AfterSales, and
MerchantSupport, respectively. It recovers 127.7\%, 166.7\%, and 39.2\% of
the F1 teacher--student gap, compared with 85.1\%, 83.3\%, and 19.6\% for
Uniform OPD. Recovery above 100\% means the student surpasses the teacher on
F1 in Fulfillment and AfterSales. The improvement on every dataset confirms
that the advantage is not confined to one business domain.

These results turn the empirical scaling relationship into an actionable
training signal. Samples with little predicted headroom provide limited new
supervision, whereas high-headroom samples expose reasoning behavior the
student has not yet mastered. Under the same budget, scaling-guided sampling
therefore concentrates optimization on higher-value instances and transfers
teacher capability more efficiently than uniform sampling.

\section{Conclusion}
\label{sec:conclusion}

This work targets compliance detection for enterprise agents with long SKILL documents. Combining formal modeling and retrieval optimization, we build the Constraint Dependency Graph to enable atomic constraint retrieval and dependency closure computation. Our approach significantly compresses the context for compliance detection while preserving detection accuracy. Evaluations over multiple datasets reveal the scaling relationship among model scale, policy complexity and detection performance. We utilize this relationship to guide sample selection in on-policy distillation and enhance the compliance detection performance of small models. Future directions include exploring diverse business scenarios, extending textual policy rules to multimodal policies, and developing techniques to further reduce compliance detection latency.

% \section*{Limitations}
% TODO: Add limitations discussion here.

\bibliography{references}

@article{li2026skillsbench,
  title={SkillsBench: Benchmarking How Well Agent Skills Work Across Diverse Tasks},
  author={Li, Xiangyi and Chen, Wenbo and Liu, Yimin and Zheng, Shenghan and Chen, Xiaokun and He, Yifeng and Li, Yubo and You, Bingran and Shen, Haotian and Sun, Jiankai and Wang, Shuyi and Li, Binxu and Zeng, Qunhong and Wang, Di and Zhao, Xuandong and Wang, Yuanli and Ben Chaim, Roey and Di, Zonglin and Gao, Yipeng and He, Junwei and He, Yizhuo and Jing, Liqiang and Kong, Luyang and Lan, Xin and Li, Jiachen and Li, Songlin and Li, Yijiang and Lin, Yueqian and Liu, Xinyi and Liu, Xuanqing and Lyu, Haoran and Ma, Ze and Wang, Bowei and Wang, Runhui and Wang, Tianyu and Ye, Wengao and Zhang, Yue and Xing, Hanwen and Xue, Yiqi and Dillmann, Steven and Lee, Han-chung},
  journal={arXiv preprint arXiv:2602.12670},
  year={2026},
  doi={10.48550/arXiv.2602.12670}
}

@misc{qwen35blog,
    title = {Qwen3.5: Accelerating Productivity with Native Multimodal Agents},
    url = {https://qwen.ai/blog?id=qwen3.5},
    author = {Qwen Team},
    month = {February},
    year = {2026}
}

@misc{deepseekv4tech,
      title={DeepSeek-V4: Towards Highly Efficient Million-Token Context Intelligence}, 
      author={DeepSeek-AI and Anyi Xu and Bangcai Lin and Bing Xue and et.al.},
      year={2026},
      eprint={2606.19348},
      archivePrefix={arXiv},
      primaryClass={cs.CL},
      url={https://arxiv.org/abs/2606.19348}, 
}

@article{han2026sweskillsbench,
  title={SWE-Skills-Bench: Do Agent Skills Actually Help in Real-World Software Engineering?},
  author={Han, Tingxu and Zhang, Yi and Song, Wei and Fang, Chunrong and Chen, Zhenyu and Sun, Youcheng and Hu, Lijie},
  journal={arXiv preprint arXiv:2603.15401},
  year={2026},
  doi={10.48550/arXiv.2603.15401}
}

@misc{zhong2026skilllearnbench,
      title={SkillLearnBench: Benchmarking Continual Learning Methods for Agent Skill Generation on Real-World Tasks}, 
      author={Shanshan Zhong and Yi Lu and Jingjie Ning and Yibing Wan and Lihan Feng and Yuyi Ao and Leonardo F. R. Ribeiro and Markus Dreyer and Sean Ammirati and Chenyan Xiong},
      year={2026},
      eprint={2604.20087},
      archivePrefix={arXiv},
      primaryClass={cs.CL},
      url={https://arxiv.org/abs/2604.20087}, 
}

@misc{gao2026skillaudit,
      title={SkillAudit: From Fixed-Suite Benchmarking to Skill-Centered Assessment}, 
      author={Dexu Yu and Youhua Li and Zhaoyang Guan and Xianhao Lin and Jining Luan and Zihao Rao and Xuanqi Lan and Yang Ran and Bo Lan and Nai-Xin Zhai and Hanwen Du and Junchen Fu and Wenhao Deng and Yongxin Ni and Chunxiao Li},
      year={2026},
      eprint={2606.22613},
      archivePrefix={arXiv},
      primaryClass={cs.AI},
      url={https://arxiv.org/abs/2606.22613}, 
}

@article{zhou2023instruction,
  title={Instruction-following evaluation for large language models},
  author={Zhou, Jeffrey and Lu, Tianjian and Mishra, Swaroop and Brahma, Siddhartha and Basu, Sujoy and Luan, Yi and Zhou, Denny and Hou, Le},
  journal={arXiv preprint arXiv:2311.07911},
  year={2023}
}

@article{he2024multi,
  title={Multi-if: Benchmarking llms on multi-turn and multilingual instructions following},
  author={He, Yun and Jin, Di and Wang, Chaoqi and Bi, Chloe and Mandyam, Karishma and Zhang, Hejia and Zhu, Chen and Li, Ning and Xu, Tengyu and Lv, Hongjiang and others},
  journal={arXiv preprint arXiv:2410.15553},
  year={2024}
}

@inproceedings{he2024can,
  title={Can large language models understand real-world complex instructions?},
  author={He, Qianyu and Zeng, Jie and Huang, Wenhao and Chen, Lina and Xiao, Jin and He, Qianxi and Zhou, Xunzhe and Liang, Jiaqing and Xiao, Yanghua},
  booktitle={Proceedings of the AAAI Conference on Artificial Intelligence},
  volume={38},
  pages={18188--18196},
  year={2024}
}

@misc{pyatkin2025generalizingverifiableinstructionfollowing,
      title={Generalizing Verifiable Instruction Following}, 
      author={Valentina Pyatkin and Saumya Malik and Victoria Graf and Hamish Ivison and Shengyi Huang and Pradeep Dasigi and Nathan Lambert and Hannaneh Hajishirzi},
      year={2025},
      eprint={2507.02833},
      archivePrefix={arXiv},
      primaryClass={cs.CL},
      url={https://arxiv.org/abs/2507.02833}, 
}

@article{wen2024benchmarking,
  title={Benchmarking complex instruction-following with multiple constraints composition},
  author={Wen, Bosi and Ke, Pei and Gu, Xiaotao and Wu, Lindong and Huang, Hao and Zhou, Jinfeng and Li, Wenchuang and Hu, Binxin and Gao, Wendy and Xu, Jiaxing and others},
  journal={Advances in Neural Information Processing Systems},
  volume={37},
  pages={137610--137645},
  year={2024}
}

@article{yao2024tau,
  title={tau-bench: A Benchmark for Tool-Agent-User Interaction in Real-World Domains},
  author={Yao, Shunyu and Shinn, Noah and Razavi, Pedram and Narasimhan, Karthik},
  journal={arXiv preprint arXiv:2406.12045},
  year={2024}
}

@article{barres2025tau,
  title={tau2-Bench: Evaluating Conversational Agents in a Dual-Control Environment},
  author={Barres, Victor and Dong, Honghua and Ray, Soham and Si, Xujie and Narasimhan, Karthik},
  journal={arXiv preprint arXiv:2506.07982},
  year={2025}
}

@article{balaji2026beyond,
  title={Beyond IVR: Benchmarking Customer Support LLM Agents for Business-Adherence},
  author={Balaji, Sumanth and Mishra, Piyush and Sachdeva, Aashraya and Agrawal, Suraj},
  journal={arXiv preprint arXiv:2601.00596},
  year={2026}
}

@misc{yang2026complibench,
      title={CompliBench: Benchmarking LLM Judges for Compliance Violation Detection in Dialogue Systems}, 
      author={Jingbo Yang and Guanyu Yao and Bairu Hou and Xinghan Yang and Nikolai Glushnev and Iwona Bialynicka-Birula and Duo Ding and Shiyu Chang},
      year={2026},
      eprint={2604.12312},
      archivePrefix={arXiv},
      primaryClass={cs.CL},
      url={https://arxiv.org/abs/2604.12312}, 
}

@article{mo2025kggen,
  title={KGGen: Extracting Knowledge Graphs from Plain Text with Language Models},
  author={Mo, Belinda and Yu, Kyssen and Kazdan, Joshua and Cabezas, Joan and Mpala, Proud and Yu, Lisa and Cundy, Chris and Kanatsoulis, Charilaos and Koyejo, Sanmi},
  journal={arXiv preprint arXiv:2502.09956},
  year={2025},
  url={https://arxiv.org/abs/2502.09956}
}

@inproceedings{snell2025scaling,
  title={Scaling LLM Test-Time Compute Optimally Can Be More Effective than Scaling Model Parameters},
  author={Snell, Charlie and Lee, Jaehoon and Xu, Kelvin and Kumar, Aviral},
  booktitle={The Thirteenth International Conference on Learning Representations},
  year={2025},
  url={https://openreview.net/forum?id=4FWAwZtd2n}
}

@inproceedings{yue2025inference,
  title={Inference Scaling for Long-Context Retrieval Augmented Generation},
  author={Yue, Zhenrui and Zhuang, Honglei and Bai, Aijun and Hui, Kai and Jagerman, Rolf and Zeng, Hansi and Qin, Zhen and Wang, Dong and Wang, Xuanhui and Bendersky, Michael},
  booktitle={The Thirteenth International Conference on Learning Representations},
  year={2025},
  url={https://arxiv.org/abs/2410.04343}
}

@article{hinton2015distilling,
  title={Distilling the Knowledge in a Neural Network},
  author={Hinton, Geoffrey and Vinyals, Oriol and Dean, Jeff},
  journal={arXiv preprint arXiv:1503.02531},
  year={2015},
  url={https://arxiv.org/abs/1503.02531}
}

@inproceedings{agarwal2024gkd,
  title={On-Policy Distillation of Language Models: Learning from Self-Generated Mistakes},
  author={Agarwal, Rishabh and Vieillard, Nino and Zhou, Yongchao and Stanczyk, Piotr and Ramos, Sabela and Geist, Matthieu and Bachem, Olivier},
  booktitle={The Twelfth International Conference on Learning Representations},
  year={2024},
  url={https://arxiv.org/abs/2306.13649}
}

@article{wang2026skilltester,
  title={SkillTester: Benchmarking Utility and Security of Agent Skills},
  author={Wang, Leye and Wang, Zixing and Xu, Anjie},
  journal={arXiv preprint arXiv:2603.28815},
  year={2026},
  url={https://arxiv.org/abs/2603.28815}
}

@article{su2026skillretrieval,
  title={Skill Retrieval Augmentation for Agentic AI},
  author={Su, Weihang and Long, Jianming and Ai, Qingyao and He, Qiaozhi and Tang, Yichen and Wang, Changyue and Tu, Yiteng and Wang, Yingbo and Liu, Yiqun},
  journal={arXiv preprint arXiv:2604.24594},
  year={2026},
  url={https://arxiv.org/abs/2604.24594}
}

@article{wu2024lifbench,
  title={LIFBench: Evaluating the Instruction Following Performance and Stability of Large Language Models in Long-Context Scenarios},
  author={Wu, Xiaodong and Wang, Minhao and Liu, Yichen and Shi, Xiaoming and Yan, He and Lu, Xiangju and Zhu, Junmin and Zhang, Wei},
  journal={arXiv preprint arXiv:2411.07037},
  year={2024},
  url={https://arxiv.org/abs/2411.07037}
}

@inproceedings{diao2025guidebench,
  title={GuideBench: Benchmarking Domain-Oriented Guideline Following for LLM Agents},
  author={Diao, Lingxiao and Xu, Xinyue and Sun, Wanxuan and Yang, Cheng and Zhang, Zhuosheng},
  booktitle={Proceedings of the 63rd Annual Meeting of the Association for Computational Linguistics},
  year={2025},
  url={https://aclanthology.org/2025.acl-long.557/}
}

@article{qi2025agentif,
  title={AGENTIF: Benchmarking Instruction Following of Large Language Models in Agentic Scenarios},
  author={Qi, Yunjia and Peng, Hao and Wang, Xiaozhi and Xin, Amy and Liu, Youfeng and Xu, Bin and Hou, Lei and Li, Juanzi},
  journal={arXiv preprint arXiv:2505.16944},
  year={2025},
  url={https://arxiv.org/abs/2505.16944}
}

@article{jiang2023longllmlingua,
  title={LongLLMLingua: Accelerating and Enhancing LLMs in Long Context Scenarios via Prompt Compression},
  author={Jiang, Huiqiang and Wu, Qianhui and Luo, Xufang and Li, Dongsheng and Lin, Chin-Yew and Yang, Yuqing and Qiu, Lili},
  journal={arXiv preprint arXiv:2310.06839},
  year={2023},
  url={https://arxiv.org/abs/2310.06839}
}

@article{pan2024llmlingua2,
  title={LLMLingua-2: Data Distillation for Efficient and Faithful Task-Agnostic Prompt Compression},
  author={Pan, Zhuoshi and Wu, Qianhui and Jiang, Huiqiang and Xia, Menglin and Luo, Xufang and Zhang, Jue and Lin, Qingwei and R{\"u}hle, Victor and Yang, Yuqing and Lin, Chin-Yew and Zhao, H. Vicky and Qiu, Lili and Zhang, Dongmei},
  journal={arXiv preprint arXiv:2403.12968},
  year={2024},
  url={https://arxiv.org/abs/2403.12968}
}

@inproceedings{du2025context,
  title={Context Length Alone Hurts LLM Performance Despite Perfect Retrieval},
  author={Du, Yufeng and Tian, Minyang and Ronanki, Srikanth and Rongali, Subendhu and Bodapati, Sravan and Galstyan, Aram and Wells, Azton and Schwartz, Roy and Huerta, Eliu A. and Peng, Hao},
  booktitle={Findings of the Association for Computational Linguistics: EMNLP 2025},
  year={2025},
  url={https://arxiv.org/abs/2510.05381}
}

@inproceedings{ouyang2022training,
  title={Training Language Models to Follow Instructions with Human Feedback},
  author={Ouyang, Long and Wu, Jeffrey and Jiang, Xu and Almeida, Diogo and Wainwright, Carroll and Mishkin, Pamela and Zhang, Chong and Agarwal, Sandhini and Slama, Katarina and Ray, Alex and Schulman, John and Hilton, Jacob and Kelton, Fraser and Miller, Luke and Simens, Maddie and Askell, Amanda and Welinder, Peter and Christiano, Paul F. and Leike, Jan and Lowe, Ryan},
  booktitle={Advances in Neural Information Processing Systems},
  volume={35},
  pages={27730--27744},
  year={2022},
  doi={10.52202/068431-2011},
  url={https://proceedings.neurips.cc/paper_files/paper/2022/hash/b1efde53be364a73914f58805a001731-Abstract.html}
}

@article{kaplan2020scaling,
  title={Scaling Laws for Neural Language Models},
  author={Kaplan, Jared and McCandlish, Sam and Henighan, Tom and Brown, Tom B. and Chess, Benjamin and Child, Rewon and Gray, Scott and Radford, Alec and Wu, Jeffrey and Amodei, Dario},
  journal={arXiv preprint arXiv:2001.08361},
  year={2020},
  doi={10.48550/arXiv.2001.08361},
  url={https://arxiv.org/abs/2001.08361}
}

@inproceedings{hoffmann2022computeoptimal,
  title={An Empirical Analysis of Compute-Optimal Large Language Model Training},
  author={Hoffmann, Jordan and Borgeaud, Sebastian and Mensch, Arthur and Buchatskaya, Elena and Cai, Trevor and Rutherford, Eliza and de Las Casas, Diego and Hendricks, Lisa Anne and Welbl, Johannes and Clark, Aidan and Hennigan, Thomas and Noland, Eric and Millican, Katherine and van den Driessche, George and Damoc, Bogdan and Guy, Aurelia and Osindero, Simon and Simonyan, Kar{\'e}n and Elsen, Erich and Vinyals, Oriol and Rae, Jack and Sifre, Laurent},
  booktitle={Advances in Neural Information Processing Systems},
  volume={35},
  pages={30016--30030},
  year={2022},
  doi={10.52202/068431-2176},
  url={https://proceedings.neurips.cc/paper_files/paper/2022/hash/c1e2faff6f588870935f114ebe04a3e5-Abstract.html}
}

@article{montgomery2025contextaware,
  title={Predicting Task Performance with Context-Aware Scaling Laws},
  author={Montgomery, Kyle and Park, David and Tu, Jianhong and Bendersky, Michael and Gunel, Beliz and Song, Dawn and Wang, Chenguang},
  journal={arXiv preprint arXiv:2510.14919},
  year={2025},
  doi={10.48550/arXiv.2510.14919},
  url={https://arxiv.org/abs/2510.14919}
}

@article{edge2024graphrag,
  title={From Local to Global: A Graph {RAG} Approach to Query-Focused Summarization},
  author={Edge, Darren and Trinh, Ha and Cheng, Newman and Bradley, Joshua and Chao, Alex and Mody, Apurva and Truitt, Steven and Metropolitansky, Dasha and Ness, Robert Osazuwa and Larson, Jonathan},
  journal={arXiv preprint arXiv:2404.16130},
  year={2024},
  doi={10.48550/arXiv.2404.16130},
  url={https://arxiv.org/abs/2404.16130}
}

@article{guo2024lightrag,
  title={{LightRAG}: Simple and Fast Retrieval-Augmented Generation},
  author={Guo, Zirui and Xia, Lianghao and Yu, Yanhua and Ao, Tu and Huang, Chao},
  journal={arXiv preprint arXiv:2410.05779},
  year={2024},
  doi={10.48550/arXiv.2410.05779},
  url={https://arxiv.org/abs/2410.05779}
}

@article{liu2024lost,
  title={Lost in the Middle: How Language Models Use Long Contexts},
  author={Liu, Nelson F. and Lin, Kevin and Hewitt, John and Paranjape, Ashwin and Bevilacqua, Michele and Petroni, Fabio and Liang, Percy},
  journal={Transactions of the Association for Computational Linguistics},
  volume={12},
  pages={157--173},
  year={2024},
  doi={10.1162/tacl_a_00638},
  url={https://aclanthology.org/2024.tacl-1.9/}
}

@inproceedings{bai2024longbench,
  title={{LongBench}: A Bilingual, Multitask Benchmark for Long Context Understanding},
  author={Bai, Yushi and Lv, Xin and Zhang, Jiajie and Lyu, Hongchang and Tang, Jiankai and Huang, Zhidian and Du, Zhengxiao and Liu, Xiao and Zeng, Aohan and Hou, Lei and Dong, Yuxiao and Tang, Jie and Li, Juanzi},
  booktitle={Proceedings of the 62nd Annual Meeting of the Association for Computational Linguistics},
  pages={3119--3137},
  year={2024},
  doi={10.18653/v1/2024.acl-long.172},
  url={https://aclanthology.org/2024.acl-long.172/}
}

@inproceedings{lewis2020rag,
  title={Retrieval-Augmented Generation for Knowledge-Intensive {NLP} Tasks},
  author={Lewis, Patrick and Perez, Ethan and Piktus, Aleksandra and Petroni, Fabio and Karpukhin, Vladimir and Goyal, Naman and K{\"u}ttler, Heinrich and Lewis, Mike and Yih, Wen-tau and Rockt{\"a}schel, Tim and Riedel, Sebastian and Kiela, Douwe},
  booktitle={Advances in Neural Information Processing Systems},
  volume={33},
  pages={9459--9474},
  year={2020},
  url={https://proceedings.neurips.cc/paper/2020/hash/6b493230205f780e1bc26945df7481e5-Abstract.html}
}

@inproceedings{gutierrez2024hipporag,
  title={{HippoRAG}: Neurobiologically Inspired Long-Term Memory for Large Language Models},
  author={Guti{\'e}rrez, Bernal Jim{\'e}nez and Shu, Yiheng and Gu, Yu and Yasunaga, Michihiro and Su, Yu},
  booktitle={Advances in Neural Information Processing Systems},
  volume={37},
  year={2024},
  doi={10.52202/079017-1902},
  url={https://proceedings.neurips.cc/paper_files/paper/2024/hash/6ddc001d07ca4f319af96a3024f6dbd1-Abstract-Conference.html}
}

@inproceedings{gu2024minillm,
  title={{MiniLLM}: On-Policy Distillation of Large Language Models},
  author={Gu, Yuxian and Dong, Li and Wei, Furu and Huang, Minlie},
  booktitle={The Twelfth International Conference on Learning Representations},
  year={2024},
  doi={10.48550/arXiv.2306.08543},
  url={https://arxiv.org/abs/2306.08543}
}

\appendix

\section{Datasets}
\label{app:data}

\subsection{Data Format}

Every sample in our datasets contains a conversation history $H$, an environment state $O$,
a candidate agent response $A$, one or more associated SKILL documents, and a
gold label $y$. Table~\ref{tab:data-schema} gives the normalized schema.
The source data use \code{positive} for a compliant response and
\code{negative} for a policy-violating response. Evaluation maps them to $y=0$
and $y=1$, respectively.

\begin{table}[htbp]
\centering
\small
\begin{tabular}{p{0.30\linewidth}p{0.52\linewidth}}
\toprule
Field & Description \\
\midrule
\code{label} & \code{positive} for compliant and \code{negative} for
violating. \\
\code{skills} & Names of the SKILL documents associated with the interaction. \\
\code{context} & Conversation history $H$ up to the audited response. \\
\code{background} & Environment state $O$, including observable business
signals. \\
\code{model\_response} & Candidate response $A$ to be audited. \\
\bottomrule
\end{tabular}
\caption{Normalized instance schema after identifier removal and
pseudonymization.}
\label{tab:data-schema}
\end{table}

The enterprise datasets originate from operational customer-service
evaluation records containing model-based assessments and an independent
human assessment. Conversion retains the positive/negative label and the
SKILL routing associated with each record. Conversation-level identifiers
become pseudonymous grouping keys. 

\subsection{Statistics}

Table~\ref{tab:data-full} reports sample counts, SKILL counts,
label distributions, and average policy lengths.
\textbf{Fulfillment} contains consumer order-fulfillment interactions governed
by multiple applicable SKILLs.
\textbf{AfterSales} contains consumer-facing post-order workflows, with one
routed SKILL per sample.
\textbf{MerchantSupport} contains merchant-facing operational-support
interactions. Together, the three datasets span complementary routing regimes,
policy structures, and service scenarios.

\begin{table}[htbp]
\centering
\small
\setlength{\tabcolsep}{1.5pt}
\begin{tabular}{lrrrr}
\toprule
Dataset & Samples  & SKILLs & Compliant & Violation  \\
\midrule
Fulfillment      & 1,442 & 14 & 955 & 487   \\
AfterSales       & 1,536 & 12 & 898 & 638    \\
MerchantSupport  &   870 & 32 & 630 & 240    \\
CompliBench      & 2,710 & 86 & 827 & 1,883  \\
CompliBench-Long & 2,710 & 86 & 827 & 1,883   \\
\bottomrule
\end{tabular}
\caption{Dataset composition and label counts. Average policy length is
measured with the Qwen3.5 series tokenizer.}
\label{tab:data-full}
\end{table}

\textbf{CompliBench.}~
We convert the public benchmark at the frozen source revision into the common
schema while preserving its conversations and labels. Each guideline becomes
one SKILL, and the original conversation identifier becomes the session
grouping key.

\textbf{CompliBench-Long.}~
This controlled variant preserves the same 2,710 conversations, 212 sessions,
86 SKILL identities, and labels as CompliBench. Domain-level distractor intents
and redundant paraphrases expand the policy text while preserving the target
decision. This paired design evaluates long-policy robustness using identical
conversations and labels.

\subsection{Sample Example}

A representative instance illustrates the normalized format:

\noindent\textbf{History $H$:} The merchant asks why a cancelled order was judged to
be the merchant's responsibility.\\
\textbf{Environment $O$:} The refund and cancellation records identify
``merchant responsibility'' and state that the item did not match its
description or was not prepared according to the note.\\
\textbf{Candidate $A$:} The agent states that the judgment may be incorrect,
promises a review within 48 hours, and redirects the merchant to a city
manager.\\
\textbf{Label:} Violation.

In this instance, $O$ supplies factual signals, $H$ establishes the decision
point, and the audit evaluates $A$ against the applicable policy obligations.

\section{CDG Construction}
\label{app:cdg}

\subsection{Offline Construction}

Each atomic constraint rule is a triple
$r_i=\langle c_i,a_i,\mathcal S_i\rangle$, where $c_i$ specifies the
applicable condition, $a_i$ defines one normative agent action (required,
prohibited, or permitted), and $\mathcal S_i$ retains the original policy text
spans for audit traceability. Algorithm~\ref{alg:construction} constructs the
rule graph block by block and then connects dependencies across blocks.

\begin{algorithm}[H]
\caption{Block-wise CDG construction for one SKILL}
\label{alg:construction}
\small
\begin{algorithmic}[1]
\REQUIRE SKILL document $D_j$ with name and description
\ENSURE Rule graph $\mathcal G_j=(\mathcal R_j,\mathcal E_j)$
\STATE Split $D_j$ at Markdown heading boundaries into semantic blocks
$B_1,\ldots,B_s$
\STATE Initialize $\mathcal G_j\leftarrow(\varnothing,\varnothing)$
\FOR{$B_\ell$ in document order}
  \STATE Extract atomic rules $\mathcal R_\ell$ from $B_\ell$
  \STATE Extract within-block \code{require} edges $\mathcal E_\ell$
  \STATE Validate fields, unique identifiers, edge endpoints, and source spans
  \STATE Rebase local rule identifiers to avoid collisions
  \IF{$\ell=1$}
    \STATE $\mathcal G_j\leftarrow(\mathcal R_\ell,\mathcal E_\ell)$
  \ELSE
    \STATE Predict cross-block merge map $\mathcal M_\ell$ and
    \code{require} edges $\mathcal E_\ell^\times$
    \STATE Apply $\mathcal M_\ell$ to rules and edge endpoints; union source spans
    \STATE Add the remaining rules in $\mathcal R_\ell$ to $\mathcal R_j$
    \STATE $\mathcal E_j\leftarrow
    \mathcal E_j\cup\mathcal E_\ell\cup\mathcal E_\ell^\times$
    \STATE Remove self-loops and duplicate edges from $\mathcal E_j$
    \STATE $\mathcal G_j\leftarrow(\mathcal R_j,\mathcal E_j)$
  \ENDIF
\ENDFOR
\STATE Store $\mathcal G_j$; index its name and description
\RETURN $\mathcal G_j$
\end{algorithmic}
\end{algorithm}

The upper layer indexes SKILL names and descriptions for scenario routing.
The lower layer models mandatory dependencies within each selected SKILL.
Rule identifiers are unique within each SKILL. Each \code{require} edge links
two existing rules and represents a mandatory audit prerequisite. A visited
set supports cyclic procedural structures and guarantees finite traversal.

\subsection{Online Inference}

Algorithm~\ref{alg:online} expands lexical seed rules along outgoing
\code{require} edges. If $r_i\rightarrow r_k$, then retrieving $r_i$ also
retrieves $r_k$. The resulting closure preserves every prerequisite represented
in the frozen graph.

\begin{algorithm}[H]
\caption{Action-conditioned compliance detection}
\label{alg:online}
\small
\begin{algorithmic}[1]
\REQUIRE Interaction $q=(H,O,A)$; two-layer CDG
\ENSURE Predicted label $\widehat y$
\STATE Extract audit keywords $K$ from $(H,O,A)$
\STATE Route to candidate SKILLs $\mathcal D_q$
\FOR{each selected SKILL $D_j\in\mathcal D_q$}
  \STATE Match $K$ against rule fields
  \STATE Let $S_j$ be the matched seed-rule identifiers
  \IF{$S_j=\varnothing$}
    \STATE $S_j\leftarrow\mathcal R_j$ for full-rule coverage
  \ENDIF
  \STATE $Q\leftarrow S_j$ and $\mathrm{Cl}_j\leftarrow S_j$
  \WHILE{$Q\neq\varnothing$}
    \STATE Remove $r_i$ from $Q$
    \FOR{each edge $(r_i,r_k)\in\mathcal E_j$}
      \IF{$r_k\notin\mathrm{Cl}_j$}
        \STATE Add $r_k$ to $\mathrm{Cl}_j$ and $Q$
      \ENDIF
    \ENDFOR
  \ENDWHILE
\ENDFOR
\STATE Render $\bigcup_j\mathrm{Cl}_j$ as condition--action text
\STATE Judge $(H,O,A)$ against the rendered closure
\RETURN $\widehat y$
\end{algorithmic}
\end{algorithm}

\subsection{Graph Statistics}

Table~\ref{tab:graph-statistics} reports the complete enterprise CDGs used in
the experiments. ``Depth'' is the mean, across SKILLs, of each graph's
maximum dependency depth. ``Cycle'' counts SKILL graphs containing at least
one directed cycle. Coupling is
$I_c=|\mathcal E|/|\mathcal R|$, computed after graph fusion. These statistics
are computed directly from the frozen policy graphs before evaluation.

\begin{table}[htbp]
\centering
\small
\setlength{\tabcolsep}{1.5pt}
\begin{tabular}{lrrrrr}
\toprule
Dataset & SKILLs & Rules & Edges & $I_c$ & Depth  \\
\midrule
Fulfillment     & 14 & 3,467 & 1,737 & 0.501 & 4.36  \\
AfterSales      & 12 & 1,235 &   869 & 0.704 & 5.33  \\
MerchantSupport & 32 & 2,494 & 1,432 & 0.574 & 3.94  \\
\midrule
Total           & 58 & 7,196 & 4,038 & 0.561 & --  \\
\bottomrule
\end{tabular}
\caption{Statistics of the constructed enterprise constraint dependency
graphs. The total coupling is computed from pooled edge and rule counts.}
\label{tab:graph-statistics}
\end{table}

The enterprise CDGs exhibit distinct structural profiles. Fulfillment has the
largest mean number of rules per SKILL ($247.6$), while MerchantSupport has
the largest SKILL set and $77.9$ rules per SKILL on average. AfterSales has the
highest dependency coupling and mean maximum depth, consistent with its staged
service workflows. Visited-set traversal handles all eight cyclic SKILL graphs
and renders each closure rule once.

\subsection{Human Evaluation}
\label{app:human}

\textbf{Gold labels.}~
Each enterprise source record includes an independent human assessment
alongside model-based assessments. Dataset conversion preserves the source
label, associated SKILLs, and session identifier. The independent assessment
supplies the gold label used throughout evaluation.

\textbf{Rule extraction.}~
Domain experts independently annotated a reference set of atomic rules for
every SKILL in the three enterprise datasets. For each dataset, all extracted
and expert-annotated rules are pooled across its SKILLs before evaluation.
Rule-set agreement measures the semantic agreement between the two pooled rule
inventories. In a separate rule-level assessment, the experts evaluated every
LLM-extracted atomic rule against its source policy; rule correctness is the
percentage of pooled extracted rules judged correct.
Table~\ref{tab:rule-evaluation} reports both dataset-level results. The two
metrics distinguish agreement with the complete expert-annotated inventory
from the correctness of individual extracted rules.

\begin{table}[htbp]
\centering
\small
\setlength{\tabcolsep}{3pt}
\begin{tabular}{lcc}
\toprule
Dataset & Rule-set agreement & Rule correctness \\
\midrule
Fulfillment     & 95.3\% & 99.8\% \\
AfterSales      & 93.7\% &  99.9\% \\
MerchantSupport & 92.8\% & 98.7\% \\
\bottomrule
\end{tabular}
\caption{Expert evaluation of LLM-extracted atomic rules on the enterprise
datasets.}
\label{tab:rule-evaluation}
\end{table}

\textbf{CDG inspection.}~
The case in Section~C is manually evaluated according to four criteria:
(i) the selected SKILL governs the interaction; (ii) the retrieved rules are
atomic and applicable; (iii) every displayed \code{require} edge expresses a
necessary audit prerequisite; and (iv) the verdict follows from the
interaction and dependency closure. The audit records a pass for all four
criteria.
The three directly applicable obligations are represented as separate atomic
actions, while dependency expansion introduces the merchant-responsibility
evidence, deduction-limit explanation, and corrective-guidance requirement
needed for the judgment. The inspection establishes an auditable path from
source policy spans to atomic rules, dependency closure, and compliance
judgment.

\subsection{Baselines}
\label{app:baselines}

\textbf{\rawskill.}~
For each sample, \rawskill concatenates the complete associated SKILL
documents in their original Markdown form. The complete policy text provides
maximal source coverage, with relevance selection and dependency reasoning
performed by the judge.

\textbf{\lightrag.}~
\lightrag builds an entity--relation index over the same SKILL corpus.
Documents are segmented into 12,000-token chunks with
100-token overlap, and all judge models evaluated on a dataset share the same
prebuilt index. Each entity and relation is profiled with textual descriptions
and linked source chunks. The online query contains the associated SKILL
names, $O$, $H$, and $A$, and requests complete condition--action constraints
for the shared compliance judge. Retrieval uses the keyword-oriented graph
query mode to collect matched entities, relations, and source chunks. The
configuration uses \code{top\_k}=20, \code{chunk\_top\_k}=8, and a
6,000-token retrieved-context cap. Online token totals cover retrieval and
judgment; index construction is part of offline preprocessing.

\textbf{\skillcdg.}~
\skillcdg selects SKILLs and audit keywords, performs lexical matching over
conditions, actions, and source spans, and computes the
transitive \code{require} closure. The rendered policy context contains rule
identifiers, conditions, and actions, while the stored graph nodes retain full
source spans for traceability.

\textbf{Ablations.}~
\emph{w/o Keyword Retrieval} provides all atomic rules in the associated
SKILLs and therefore removes both keyword extraction and two-level filtering.
\emph{w/o Closure} retains SKILL selection and lexical seed matching and passes
the matched seed rules directly to the judge. These variants isolate context
filtering from dependency completion.

\begin{table}[htbp]
\centering
\tiny
\begin{tabular}{lccc}
\toprule
Method & Policy unit & Filtering & Dependency expansion \\
\midrule
\rawskill & Raw documents & None & Implicit in judge \\
\lightrag & Retrieved text & Keyword graph & None \\
\skillcdg & Atomic rules & Two-level & Transitive \code{require} \\
w/o Keyword & Atomic rules & None & All rules supplied \\
w/o Closure & Atomic rules & Two-level & None \\
\bottomrule
\end{tabular}
\caption{Policy-access methods and controlled ablations.}
\label{tab:methods}
\end{table}

\textbf{Shared controls.}~
For every comparison, the compliance-judgment prompt, response schema,
decoding settings, candidate responses, and labels are fixed. The response
schema requires a Boolean \code{violation} and a short reason. Each result uses
one completed judgment per instance. The parser validates Boolean labels and
accompanying reasons before aggregation. Online token accounting includes
selection and judgment calls; index and CDG construction belong to offline
preprocessing.

\section{Case Study}
\label{app:case}

We examine a MerchantSupport violation instance using DeepSeek-V4-Flash as the
fixed judge.

\subsection{Interaction}

\begin{quote}\small
\textbf{Conversation.}~
The merchant asks why an order was judged to violate the rules. In the prior
turn, the agent explains that the cancellation note says ``item inconsistent
with the description / not prepared according to the order note.'' The
merchant then objects that bundled-discount orders generally do not support
customer notes.

\textbf{Relevant environment state.}~
Both the refund record and the cancellation record assign responsibility to
the merchant. The cancellation evidence states that the item was inconsistent
with its description or not prepared according to the order note. The customer
refund is 11.16 currency units, and the merchant's expected income, which
bounds the relevant deduction explanation, is 8.66.

\textbf{Candidate response.}~
``You are right that bundled-discount orders do not support customer notes.
The basis for this judgment may be incorrect. I will record it for further
review, and a result is expected within 48 hours. For the meal-loss claim,
please contact the city manager. Would you like me to find the contact
information?''
\end{quote}

The response discounts the recorded cancellation evidence, omits the
deduction limit and corrective recommendation, and introduces an unsupported
48-hour review commitment.

\subsection{Extracted CDG}

The graph-construction model extracts six relevant rules and three
\code{require} edges from the order-refund SKILL. The complete
\code{sources} arrays are retained in the stored rule nodes.

\begin{lstlisting}[style=prompt]
{
  "rules": [
    {"id": "C032",
     "condition": "merchant responsibility and refund explanation",
     "action": "state the refund reason and cite responsibility evidence"},
    {"id": "C033",
     "condition": "merchant responsibility and refund explanation",
     "action": "state the deduction upper bound"},
    {"id": "C034",
     "condition": "merchant responsibility and refund explanation",
     "action": "provide an improvement recommendation"},
    {"id": "C014",
     "condition": "determining the refund reason or responsible party",
     "action": "read both from system signals"},
    {"id": "C050",
     "condition": "using the merchant-responsibility refund script",
     "action": "state merchant responsibility and the deduction limit"},
    {"id": "C007",
     "condition": "the refund is caused by the merchant",
     "action": "explain honestly and provide corrective guidance"}
  ],
  "edges": [
    {"source": "C032", "target": "C014", "type": "require"},
    {"source": "C033", "target": "C050", "type": "require"},
    {"source": "C034", "target": "C007", "type": "require"}
  ]
}
\end{lstlisting}

Rules \code{C032}--\code{C034} are directly applicable to the interaction.
Their dependency closure adds \code{C014}, \code{C050}, and \code{C007},
thereby grounding the required evidence, deduction explanation, and
corrective guidance. Every rendered business fact is grounded in the source
policy.

\subsection{Detection Output}

For each policy representation, we report the fixed judge's Boolean
\code{violation} decision and reason.

\textbf{\rawskill.}~
\begin{lstlisting}[style=prompt]
{
  "violation": false,
  "reason": "The response explains the refund and provides a city-manager route without exceeding the SKILL scope."
}
\end{lstlisting}

\textbf{\skillcdg.}~
\begin{lstlisting}[style=prompt]
{
  "violation": true,
  "reason": "Rules C033-C034 require the 8.66 limit and improvement advice; the response gives review and city-manager routing."
}
\end{lstlisting}

The contrast centers on policy structure: \rawskill presents the policy as a
single text sequence, whereas \skillcdg renders the three obligations as
separate atomic rules and connects them to their prerequisites through
dependency closure. 
The frozen CDG maps each rule identifier to its condition,
action, and source span, making retrieval, dependency completion, and rule
application directly inspectable.

\section{Training}
\label{app:training}

\subsection{Data}

The two OPD variants draw training instances from the same three enterprise
datasets used in the main experiments. Each compliance-judgment input follows
the same \skillcdg construction and contains the conversation history $H$,
environment state $O$, candidate response $A$, and retrieved policy context.
The inputs are materialized before sample selection so that both methods use
the same fixed representation. Direct identifiers are replaced, and sample
and session identifiers are stored as irreversible SHA-256 pseudonyms.

Each method trains on 2,000 selected instances under identical dataset--label
quotas. Uniform OPD samples randomly within each stratum using seed 17.
Scaling-guided OPD ranks instances within the same stratum by
$g_i=\widehat P_{9\mathrm{B}}(C_i,d_i)-
\widehat P_{4\mathrm{B}}(C_i,d_i)$ and retains the highest-scoring instances.
This matched design fixes the training-set size and composition across the two
allocation strategies.

\subsection{Environment}

Table~\ref{tab:compute-environment} summarizes the execution environments.
Open-source judges in the main detection experiments are served with SGLang
on eight NVIDIA H20 GPUs with 96\,GB memory per GPU; closed-source judges are
accessed through their APIs. OPD is executed on two NVIDIA RTX PRO 6000 GPUs
with 96\,GB memory per GPU: one GPU hosts student training and rollouts, and
one hosts teacher inference. Student rollouts and teacher inference use vLLM,
while verl coordinates on-policy generation and the policy-gradient update.

\begin{table}[htbp]
\centering
\small
\begin{tabular}{p{0.25\linewidth}p{0.6\linewidth}}
\toprule
Component & Configuration \\
\midrule
Main inference &
$8\times$ NVIDIA H20 (96\,GB); SGLang for open-source models; API inference
for closed-source models. \\
Main decoding &
Temperature 0.01 with a fixed prompt and response schema across methods. \\
OPD hardware &
$2\times$ NVIDIA RTX PRO 6000 (96\,GB), Ubuntu 22.04. \\
OPD software &
Python 3.10, PyTorch 2.10, CUDA 12.8, verl 0.8.0, and vLLM 0.18.0. \\
Models &
Qwen3.5-4B student and Qwen3.5-9B teacher, loaded from local checkpoints. \\
Precision &
BF16 student optimization with FSDP and gradient checkpointing. \\
\bottomrule
\end{tabular}
\caption{Compute environments used for evaluation and OPD.}
\label{tab:compute-environment}
\end{table}

Each reported model--method--dataset cell is derived from a complete
end-to-end execution. RQ2 uncertainty is quantified with 1,000 session-clustered
bootstrap replicates using seed 2027. OPD sample selection and training use
seed 17.

\subsection{Optimization}

Table~\ref{tab:opd-hparams} lists the shared configuration for Uniform OPD and
Scaling-guided OPD. Each run processes 2,000 instances for two epochs with a
global batch size of 24, yielding 168 optimizer steps. The student is updated
through rank-32 LoRA adapters on the language-model linear layers; the visual
modules, \code{lm\_head}, and \code{embed\_tokens} remain frozen. The model
pair, optimization schedule, and decoding settings are shared across the two
methods.

\begin{table}[htbp]
\centering
\small
\begin{tabular}{p{0.38\linewidth}p{0.5\linewidth}}
\toprule
Parameter & Value \\
\midrule
Teacher / student & Qwen3.5-9B / Qwen3.5-4B \\
Training instances & 2,000 \\
Epochs / optimizer steps & 2 / 168 \\
Global batch size & 24 \\
Policy mini-batch size & 24 \\
Learning rate & $1\times10^{-6}$ \\
LoRA rank / alpha & 32 / 64 \\
Maximum prompt length & 12,288 tokens \\
Maximum response length & 128 tokens \\
Maximum tokens per GPU & 24,834 \\
Rollouts per prompt $n$ & 1 \\
Rollout temperature / top-$p$ & 0.7 / 0.9 \\
KL estimator & Single-sample $k_1$ reverse-KL estimator \\
Policy update & Vanilla policy gradient with detached $-k_1$ advantages \\
Task reward & Disabled \\
$k_1$ clamp & $[-10,10]$ \\
Training seed & 17 \\
\bottomrule
\end{tabular}
\caption{Shared OPD hyperparameters.}
\label{tab:opd-hparams}
\end{table}

For each training prompt, the student samples one response from its current
policy. At each response token $a_t$, the frozen teacher scores the same token
conditioned on the student-generated prefix $x_{<t}$. The verl trainer computes
$k_{1,t}=\log\pi_\theta(a_t\mid x_{<t})-
\log\pi_T(a_t\mid x_{<t})$; its expectation under
$a_t\sim\pi_\theta(\cdot\mid x_{<t})$ is
$D_{\mathrm{KL}}(\pi_\theta\,\|\,\pi_T)$. The estimator is clipped, negated,
and used as a detached advantage in the policy-gradient loss. With task rewards
disabled, teacher-derived advantages provide the optimization signal on
trajectories drawn from the current student policy.

\end{document}